\documentclass[times, review, 10pt]{elsarticle}
\usepackage{amssymb}
\usepackage{graphicx}
\usepackage{amsmath}
\usepackage{longtable}
\usepackage[letterpaper,
top=4.3cm,
bottom=4.3cm,
left=4.8cm,
right=4.8cm]{geometry}
\usepackage{mathrsfs}
\usepackage{subcaption}
\usepackage{mathtools}
\usepackage{pifont}
\usepackage{algorithmic}
\usepackage{algorithm}
\usepackage{xcolor}
\usepackage[T1]{fontenc}
\usepackage{color}
\usepackage{lineno}
\usepackage{makecell} 
\usepackage{pdflscape}
\usepackage{adjustbox}
\usepackage[utf8]{inputenc}
\usepackage{tabularx}
\usepackage{setspace}
\usepackage{blindtext}
\usepackage{multirow}
\usepackage{notoccite}
\usepackage{lscape}
\usepackage{caption}
\usepackage{mwe}
\usepackage{tikz}
\usepackage{siunitx}
\usepackage{mathrsfs}
\usetikzlibrary{shapes,arrows}
\usepackage{xcolor}
\usepackage{booktabs}
\usepackage{hyperref}
\usepackage{amsthm}

\newcommand{\RNum}[1]{\lowercase\expandafter{\romannumeral #1\relax}}
\newcommand{\RNumU}[1]{\uppercase\expandafter{\romannumeral #1\relax}}
\usepackage{natbib}

\journal{Pattern Recognition}

\begin{document}
\begin{frontmatter}
\title{GBTSVR: Twin Support Vector Regression via Granular Ball Computing}
\author[inst1]{A. Quadir}
\ead{mscphd2207141002@iiti.ac.in}
\author[inst1]{M. Tanveer\corref{Correspondingauthor}}
\ead{mtanveer@iiti.ac.in}

\affiliation[inst1]{organization={Department of Mathematics, Indian Institute of Technology Indore},
            addressline={Simrol}, 
            city={Indore},
            postcode={453552}, 
            state={Madhya Pradesh},
            country={India}}
            \cortext[Correspondingauthor]{Corresponding author}
            
\begin{abstract}
Regression models used in real-world applications are often challenged by noise, outliers, data heterogeneity, and asymmetric error characteristics, which severely limit the effectiveness of conventional point-wise learning models. While twin support vector regression (TSVR) partially addresses these issues through the use of nonparallel regression bounds, it still relies on individual samples as basic learning units. As a result, TSVR remains highly sensitive to noisy and contaminated observations and fails to leverage the underlying structural and regional information inherent in complex datasets, leading to suboptimal robustness and generalization performance. To address these limitations, we introduce granular ball twin support vector regression (GBTSVR) and its quantile extension, granular ball twin support vector quantile regression (GB-TSVQR). The proposed models integrate granular ball computing with TSVR, replacing individual data points with compact, structure-preserving granular balls as fundamental learning units. This region-level representation effectively suppresses noise and outliers while preserving intrinsic data geometry, thereby enhancing robustness, stability, and generalization. GBTSVR constructs nonparallel lower- and upper-bound regression functions over granular balls, significantly reducing sensitivity to sample-level perturbations. Building upon this framework, GB-TSVQR further incorporates quantile regression principles to simultaneously capture heterogeneous trends and asymmetric error distributions across different quantile levels. Extensive experiments on synthetic datasets, time-series benchmarks, and outlier-contaminated data demonstrate that the proposed GBTSVR and GB-TSVQR consistently outperform state-of-the-art regression models. The results confirm that learning at an appropriate granularity level, combined with twin and quantile regression principles, provides a powerful and scalable solution for complex regression tasks.
\end{abstract}
\end{frontmatter}

\section{Introduction}
Machine learning, as an integral branch of artificial intelligence, has undergone rapid development and has become a transformative force across numerous scientific and engineering disciplines due to its capability to uncover hidden structures and patterns from complex datasets \cite{goldenberg2019new}. Within this broad area, supervised learning plays a pivotal role, where models are trained using labeled data to establish relationships between input features and corresponding outputs, enabling accurate prediction and inference \cite{hastie2008overview, liu2026projection}. Regression, in particular, constitutes a core task in supervised learning, focusing on learning functional mappings between explanatory variables and continuous response values \cite{11353218}.

Among the various regression techniques, support vector regression (SVR) \cite{basak2007support} has gained considerable attention owing to its strong theoretical foundation and robust predictive performance. SVR is derived from the support vector machine (SVM) framework \cite{cortes1995support}, which is widely recognized for its success in classification problems. SVM-based models have demonstrated effectiveness in a wide range of applications, including medical diagnosis such as Alzheimer’s diagnosis \cite{10619989}, EEG signal classification \cite{richhariya2018eeg} among many others. Grounded in statistical learning theory, SVMs are designed to achieve good generalization performance when applied to unseen data \cite{quadir2024intuitionistic}. By extending the SVM formulation to accommodate continuous outputs, SVR becomes a powerful tool for regression and function approximation tasks \cite{wang2023new}. Consequently, SVR has been successfully applied in diverse real-world scenarios, including brain age estimation \cite{tanveer2023deep, QUADIR2026108449}, long-term stock index forecasting \cite{5406166}, prediction of gas diffusion coefficients \cite{onyekwena2022support}, bank profitability estimation \cite{zahariev2022estimation}, and financial market analysis \cite{zhang2022application}.

However, most conventional SVR-based approaches construct the decision function using two parallel hyperplanes, which may fail to adequately capture the inherent asymmetry present in real-world data distributions. This limitation can lead to the loss of valuable asymmetric information associated with different data points. To overcome this drawback, twin support vector regression (TSVR) \cite{peng2010tsvr} is introduced, which learns two nonparallel regression functions simultaneously. These functions are designed to estimate the upper and lower $\varepsilon$-insensitive bounds of the unknown regression function, thereby effectively modeling asymmetric data characteristics and providing a more flexible and informative representation of the underlying regression structure. \citet{shao2013varepsilon} enhanced the TSVR framework by incorporating a regularization term, leading to the development of the $\varepsilon$-twin support vector regression ($\varepsilon$-TSVR). This formulation results in stable, positive definite quadratic programming problems in the dual space and significantly improves regression performance. To further enhance generalization ability, \citet{xu2018asymmetric} proposed an asymmetric $\nu$-twin support vector regression with pinball loss (Asy-$\nu$-TSVR), which explicitly accounts for data asymmetry. Later, \citet{gupta2021robust} introduced the asymmetric Lagrangian $\nu$-twin support vector regression with pinball loss (URALTSVR), enabling finer control over fitting errors within an asymmetric tube. 
Although the TSVR framework, through its two nonparallel hyperplanes, effectively captures asymmetric characteristics in data, it may still overlook latent heterogeneous information that is not directly observable across all data points. To simultaneously model both unobserved heterogeneity and asymmetric behavior, the concept of quantile regression (QR) is integrated into the $\varepsilon$-TSVR framework, giving rise to the twin support vector quantile regression (TSVQR) model \cite{ye2024twin}. This integration allows the regression model to provide a more comprehensive and flexible representation of the conditional distribution of the response variable. Despite significant progress in SVR and its variants, existing regression frameworks continue to face persistent challenges when applied to real-world data. In particular, most regression models operate at the level of individual data points, making them inherently sensitive to noise, outliers, and sample redundancy. This point-wise learning paradigm not only degrades predictive stability under data contamination but also leads to high computational overhead, especially for large-scale and noisy datasets. Moreover, classical TSVR-based models, while capable of capturing asymmetric error structures through dual nonparallel regressors, still rely on raw samples and therefore remain vulnerable to heterogeneity and local irregularities in the data distribution \cite{tanveer2024comprehensive}. 

Human cognition typically follows a global-to-local perception strategy, forming a coarse overall understanding before refining details \cite{liu2026topological}. In visual perception, the brain first captures global structure and topology, then focuses on local variations. In contrast, many machine learning models learn at the level of individual samples or features, which can lead to issues with scalability, computational cost, and robustness in large, noisy, or complex regression datasets.

Granular computing addresses this by organizing data into higher-level information granules, enabling learning at appropriate abstraction levels. Granular ball (GB) learning, introduced by \citet{xia2019granular}, partitions data into hyperspherical regions of varying radii, where each granular ball represents a local data distribution and preserves key structural characteristics while reducing noise \cite{9139397}. Prior work shows that selecting suitable granularity improves generalization and scalability, aligning with hierarchical human information processing \cite{xia2021granular}.

Granular computing provides an alternative perspective by learning at suitable abstraction levels rather than processing individual samples \cite{xia2022efficient}. It groups data into meaningful granules that retain intrinsic structure while reducing irrelevant variation. Among these methods, granular ball computing represents data using compact, high-purity hyperspherical regions, effectively aggregating similar samples to reduce redundancy and noise while capturing underlying distributions \cite{10619989}. Although granular ball–based approaches have shown strong classification performance \cite{10759815, tanveer2025enhancing}, their application to regression, particularly within the TSVR framework, remains limited, motivating this work \cite{quadir2024enhanced}.

To address the limitations of point-wise regression learning, we propose granular ball twin support vector regression (GBTSVR), a novel regression framework that integrates granular ball computing with TSVR. In GBTSVR, granular balls are treated as the fundamental learning units instead of individual samples. Each granular ball encapsulates local data geometry, enabling the regression model to learn from structured regions rather than isolated points. This design significantly reduces sensitivity to noise and outliers while preserving essential regression trends. Building upon this framework, we further introduce granular ball twin support vector quantile regression (GB-TSVQR), which incorporates quantile regression into the granular ball–based TSVR architecture. By jointly modeling lower and upper nonparallel quantile regressors at different quantile levels, GB-TSVQR captures both heterogeneous and asymmetric information in the data distribution. The use of granular balls ensures that quantile estimation is robust to extreme deviations and local distributional irregularities, while the twin-regressor formulation maintains computational efficiency through smaller, well-structured optimization problems.

The main contributions of this paper are summarized as follows:
\begin{itemize}
    \item We propose granular ball twin support vector regression (GBTSVR), a novel regression framework that integrates granular ball computing with twin support vector regression. By using granular balls as region-level learning units instead of individual samples, the proposed model effectively mitigates noise, reduces redundancy, and enhances robustness in regression tasks.
    \item We further introduce granular ball twin support vector quantile regression (GB-TSVQR), which extends GBTSVR by incorporating quantile regression. This model jointly captures heterogeneous and asymmetric distributional characteristics across different quantile levels, providing a more comprehensive representation of the underlying data structure.
    \item The proposed GB-TSVQR formulates the learning task as two reduced-size quadratic programming problems (QPP), each associated with a single set of constraints defined over granular balls rather. Furthermore, a dual coordinate descent algorithm is employed to solve each QPP efficiently, resulting in significantly accelerated training while maintaining high regression accuracy.
    \item Comprehensive experiments on synthetic datasets, time-series datasets, and outlier-injected regression benchmarks demonstrate that GBTSVR and GB-TSVQR consistently outperform state-of-the-art regression models in terms of accuracy, robustness, and stability across diverse noise settings.
\end{itemize}
Related work on granular-ball computing and the mathematical formulation of TSVR are presented in Section S.I of the Supplementary Material.
\begin{figure*}[ht!]
\begin{minipage}{.30\linewidth}
\centering
\subfloat[The original \\ dataset]{\label{G1a}\includegraphics[scale=0.16]{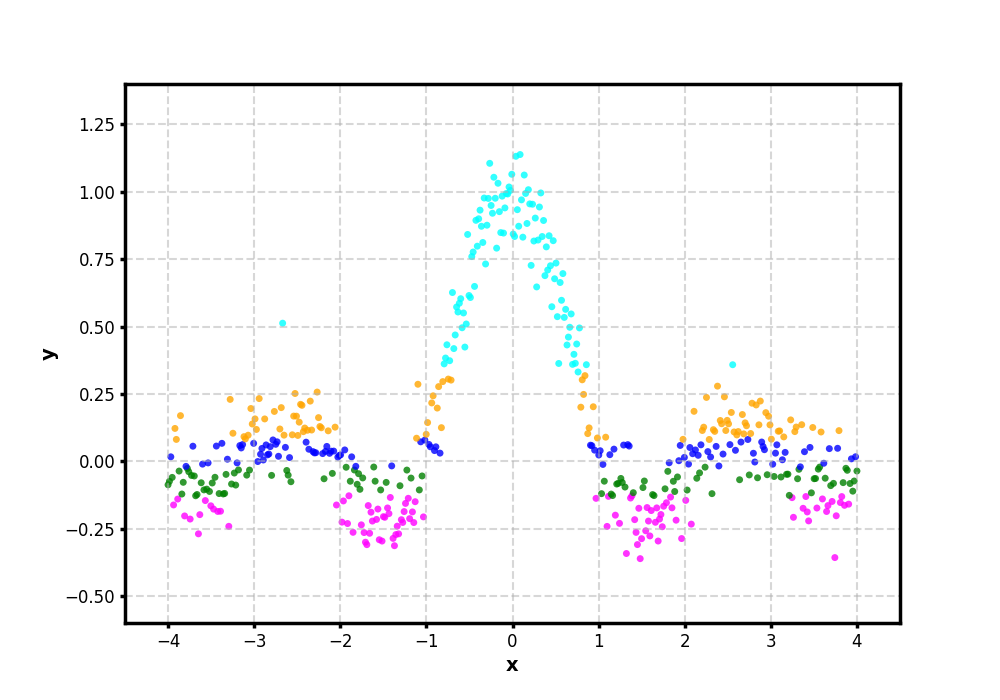}}
\end{minipage}
\begin{minipage}{.30\linewidth}
\centering
\subfloat[Generated granular ball in first \\ iteration]{\label{G1b}\includegraphics[scale=0.16]{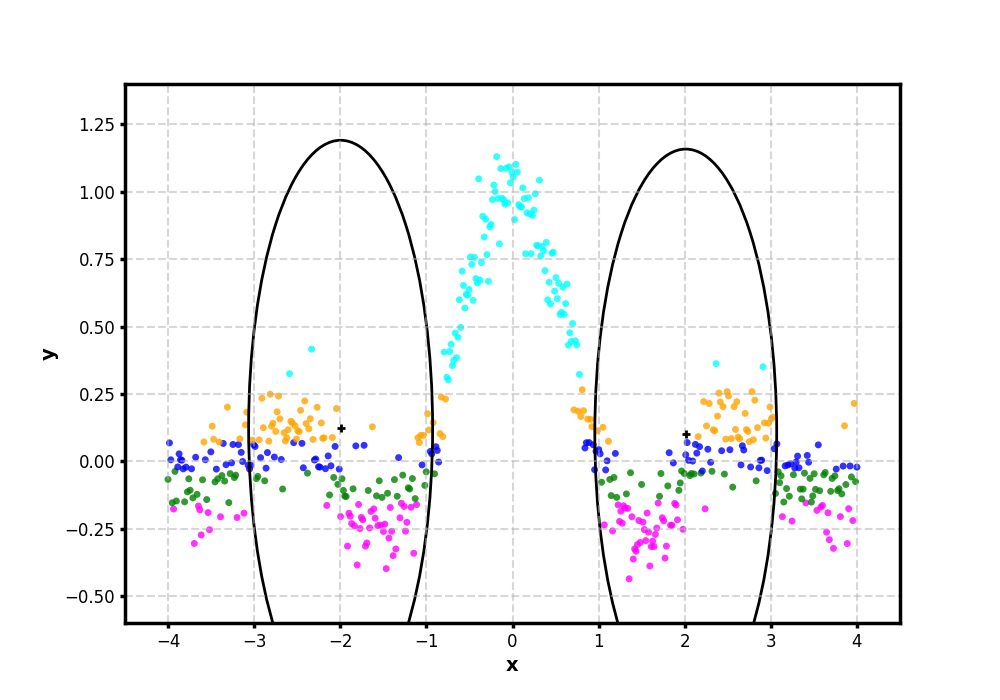}}
\end{minipage}
\begin{minipage}{.30\linewidth}
\centering
\subfloat[Generated granular ball in second iteration]{\label{G1c}\includegraphics[scale=0.16]{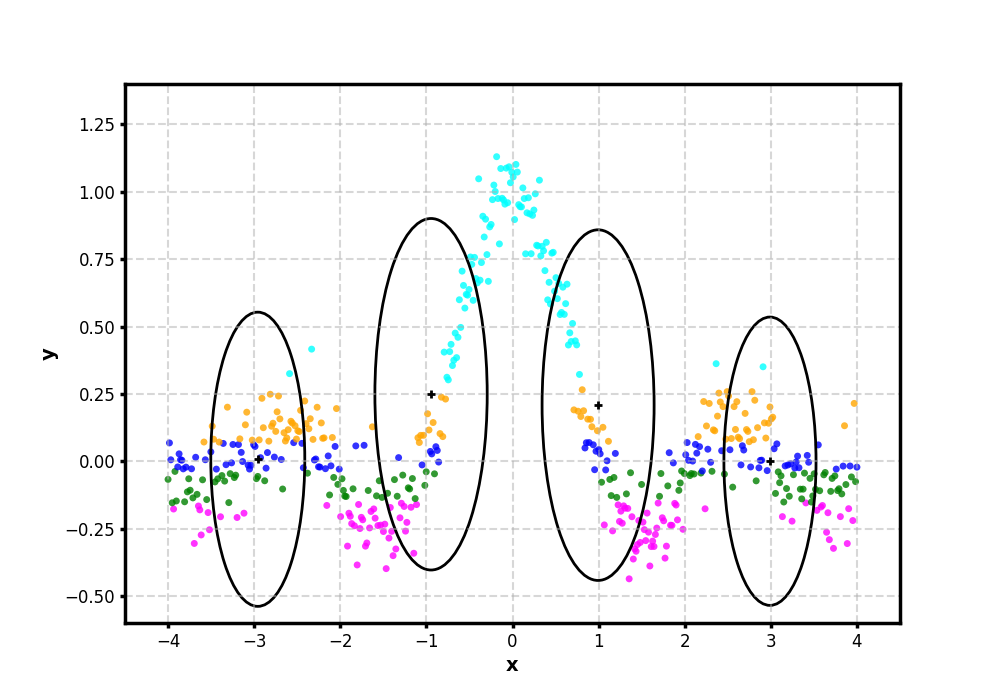}}
\end{minipage}
\par\medskip
\par\medskip
\begin{minipage}{.30\linewidth}
\centering
\subfloat[Generated granular ball in middle iteration]{\label{G1d}\includegraphics[scale=0.16]{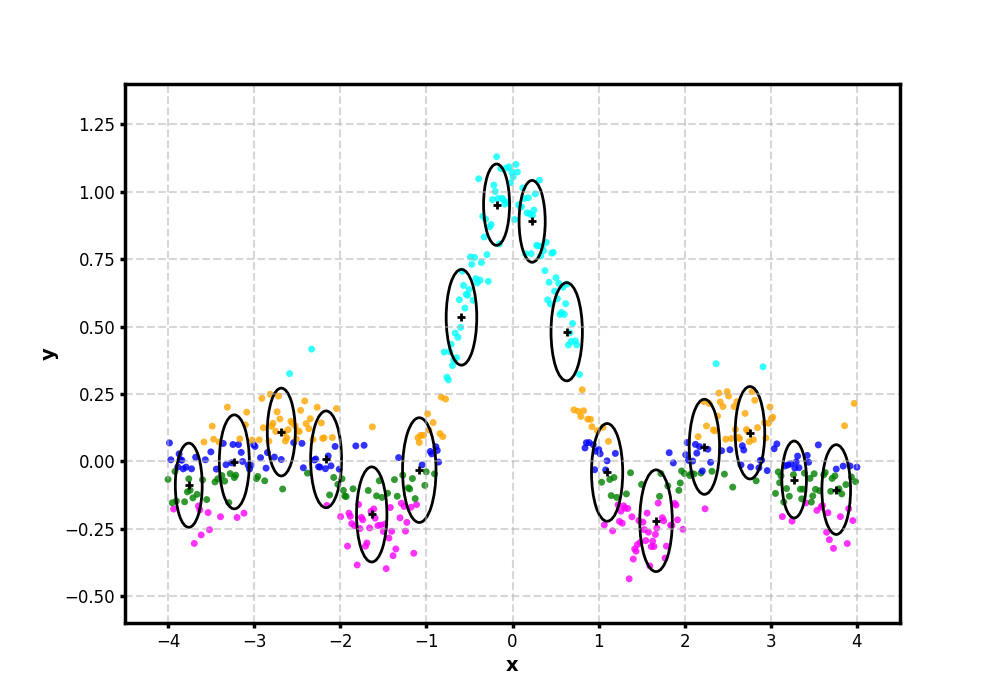}}
\end{minipage}
\begin{minipage}{.30\linewidth}
\centering
\subfloat[Results after stop \\ splitting]{\label{G1e}\includegraphics[scale=0.16]{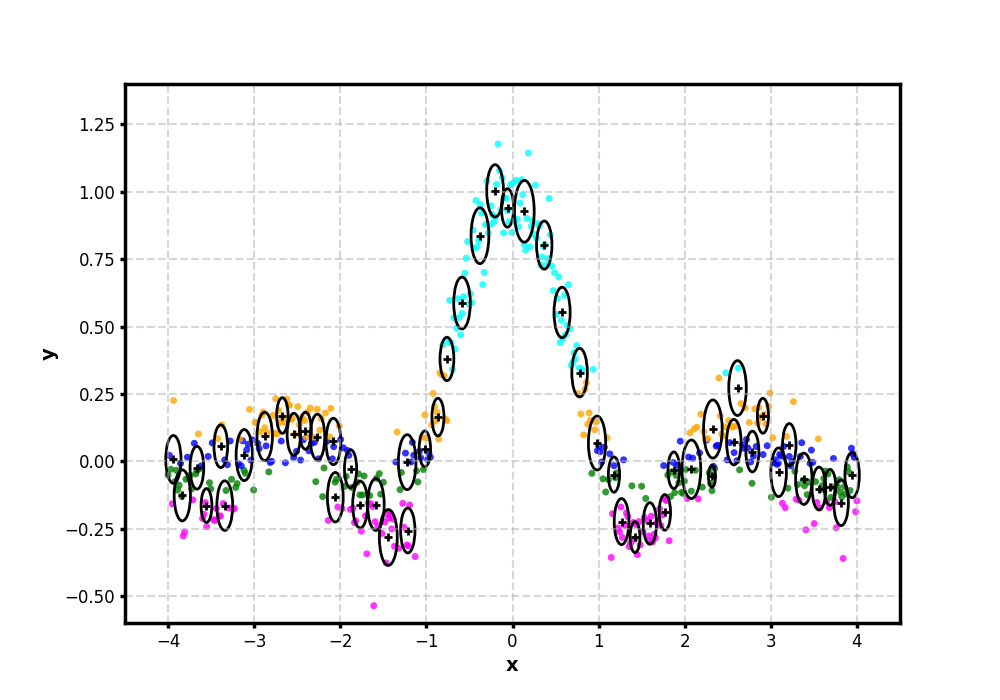}}
\end{minipage}
\begin{minipage}{.30\linewidth}
\centering
\subfloat[Extracted granular \\ balls]{\label{G1f}\includegraphics[scale=0.16]{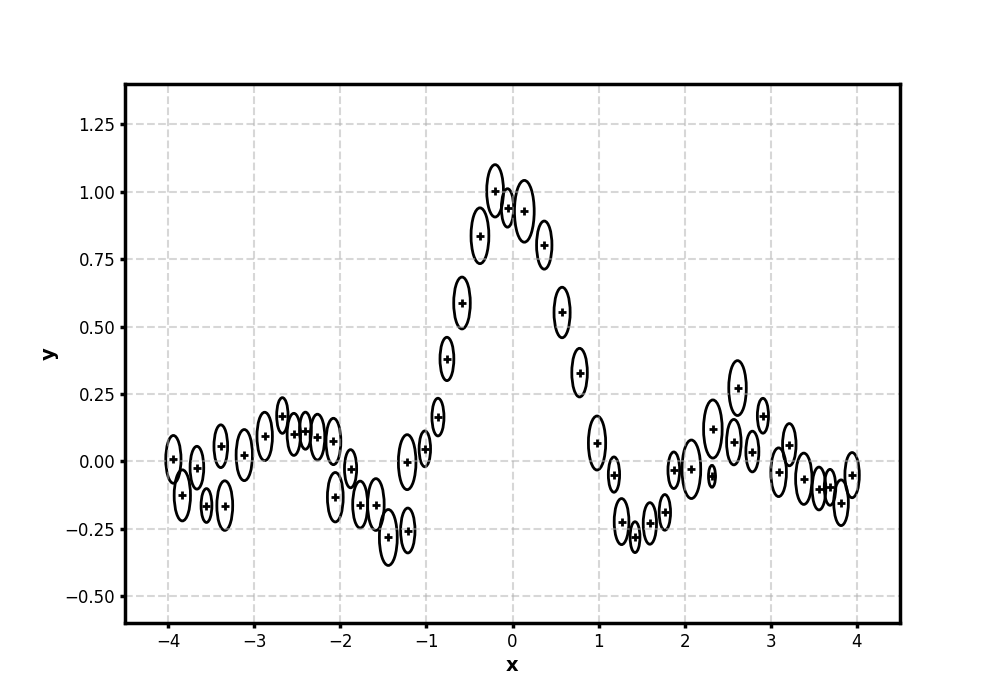}}
\end{minipage}
\caption{This figure demonstrates the construction of granular balls by segmenting the sinc dataset corrupted with Gaussian noise of zero mean and a standard deviation of $0.2$.}
\label{Gen Granular ball}
\end{figure*}

\begin{figure}[ht!]
    \centering
    \includegraphics[width=0.6\textwidth,height=6cm]{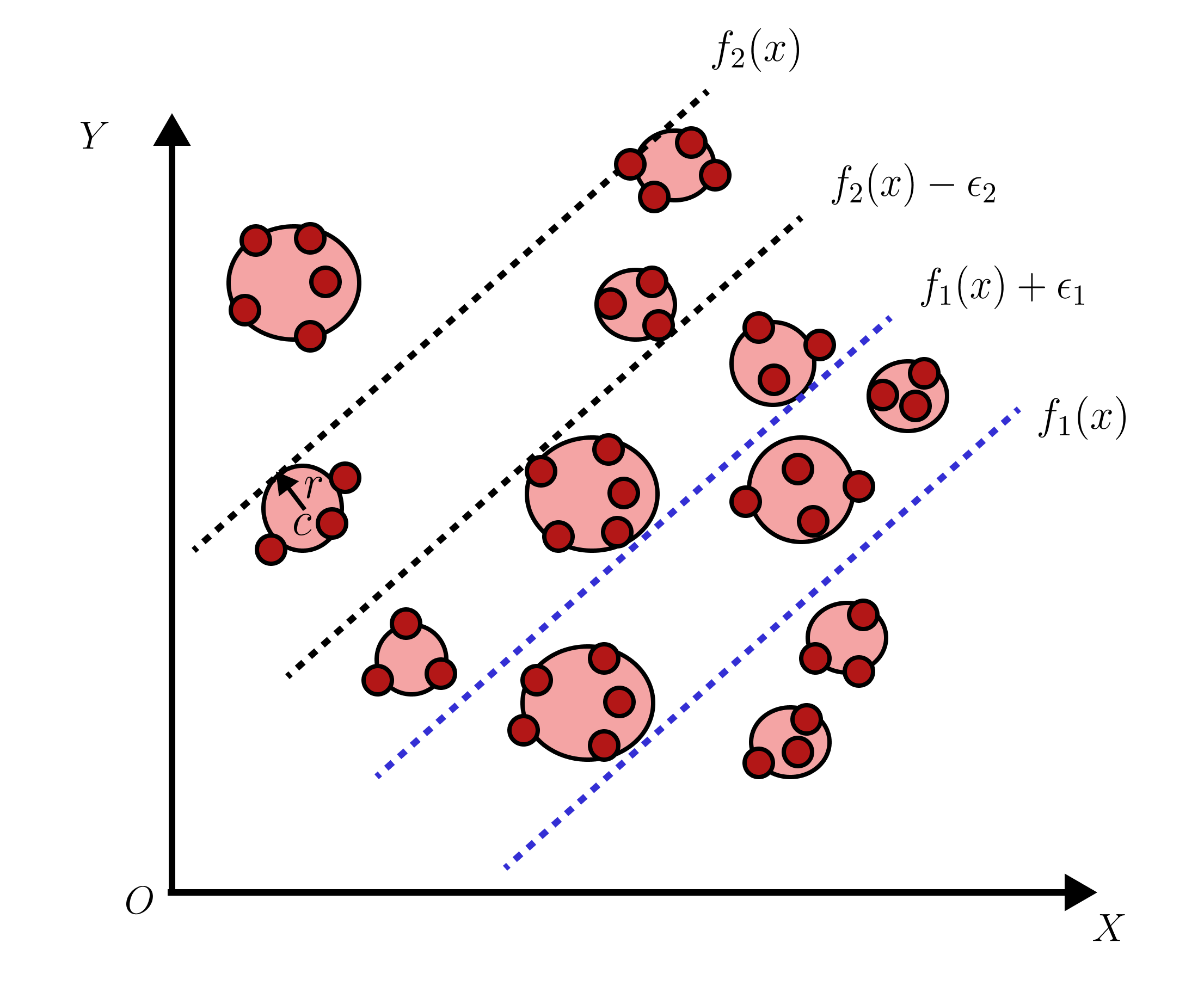}
    \caption{Overview of the GBTSVR Framework.}
    \label{Geometrical structure}
\end{figure}

\section{Granular-Ball–Based Regression Framework}
Let a regression dataset be denoted by $\mathcal{D} = \{(x_i, t_i)\}_{i=1}^n$, $x_i \in \mathbb{R}^l$, $t_i \in \mathbb{R} \}$, where $x_i$ represents the input vector and $y_i$ is the corresponding response. Directly learning regression functions from point-wise samples is often vulnerable to noise, outliers, and redundant observations. To address these limitations, the proposed framework first transforms the original dataset into a compact set of structured granules and subsequently performs regression at the granule level. 

To enable the use of purity concepts originally defined for classification, the continuous response variable is first discretized into a finite set of categories. Specifically, the target variable $t \in \mathbb{R}$ is partitioned into $k$ non-overlapping intervals, where each interval groups samples exhibiting similar response characteristics. All samples falling within the same interval are assigned a common discrete label from $\{1,2, \ldots, k\}$. This discretization step allows the notion of purity to be extended to regression settings and serves as the foundation for defining a regression-oriented granularity quality measure.

A granular ball $GB(c, r)$ summarizes a subset of data points through its center $c \in \mathbb{R}^l$ and radius $r \in \mathbb{R^+}$, and is defined as $GB(c, r) = \{x \in \mathbb{R}^l | ~ d(x, c) \leq r\}$, where $d(x, c) = \sqrt{\sum_{i=1}^l (x_i -c_i)^2}$ denotes the Euclidean distance. For a granular ball containing $p$ samples $\{x_j\}_{j=1}^p$, the center is computed as the centroid of the enclosed points $c = \frac{1}{p}\sum_{j=1}^px_j$. The radius quantifies the dispersion of samples within the ball and is defined using the average distance to improve robustness against outliers and is defined as $r = \frac{1}{p} \sum_{j = 1}^p \|x_j - c\|$. Each granular ball thus captures both the central tendency and the local variability of a group of neighboring samples, providing a compact and noise-tolerant data representation. 

To ensure that each granular ball corresponds to a coherent local structure, a purity criterion is imposed. For a granular ball $GB_i$ purity is defined as the proportion of samples sharing the dominant label within the ball. A predefined threshold $\tau \in (0, 1]$ enforces homogeneity among the enclosed samples. Let $\{GB_i\}_{i = 1}^p$ denote the set of granular balls obtained after partitioning the dataset. The generation of granular balls is formulated as the following optimization problem: 
\begin{align}
&\min~\varrho_1 \cdot \frac{n}{\sum_{i=1}^p |GB_i|} + \varrho_2 \cdot p \nonumber \\
& \text{s.t.}~~ \text{purity}(GB_j) \geq \tau, ~~j = 1, 2, \ldots, p, 
\end{align}
where $n$ is the total number of samples, $|GB_i|$ dnotes the cardinality of the $i^{th}$ granular ball, $p$ is the number of granular balls, and $\varrho_1$, $\varrho_2$ are weighting coefficients that balance compactness and granularity. The construction process follows an iterative coarse-to-fine strategy. Initially, all samples are grouped into a single granular ball. If its purity is below the threshold $\tau$, the ball is split into smaller granules. This procedure continues until all granular balls satisfy the purity constraint, yielding a set of compact and informative granules that preserve the intrinsic structure of the data.

At the outset, the entire training dataset is treated as a single granular ball, as shown in Figure~\ref{G1a}. This initial granule is then iteratively partitioned based on the proposed regression-oriented quality criterion, leading to progressively more homogeneous granular balls, as illustrated in Figures~\ref{G1b}–\ref{G1d}. The splitting process continues until all granular balls satisfy the predefined regression purity threshold, at which point the algorithm converges, as depicted in Figure~\ref{G1e}. The final set of granular balls, shown in Figure~\ref{G1f}, demonstrates the capability of granular ball computing to effectively capture the intrinsic structure and distributional characteristics of the regression data.

\section{The Proposed Granular Ball Twin Support Vector Regression}
In this section, we present the mathematical formulation of the proposed granular ball twin support vector regression (GBTSVR) model. The core idea of GBTSVR is to integrate granular ball computing with twin support vector regression by constructing regression functions using granular balls as learning units instead of individual data samples. Each granular ball summarizes a local region of the input space through its center, radius, and associated statistical information, enabling the model to capture the intrinsic data structure while suppressing the influence of noise and outliers. By learning two nonparallel regression functions from these granular representations, GBTSVR achieves improved robustness, enhanced generalization, and reduced computational complexity. The mathematical formulation of GBTSVR is given as follows:
\begin{align}
\label{eq:1}
    \underset{w_1, b_1}{\min} ~& \frac{1}{2} \|\mathcal{Y} + \mathcal{R} - e \epsilon_1 - (\mathcal{C}w_1 + eb_1)\|^2 + \mathcal{D}_1 e^T \eta_1 \nonumber \\
    \text{s.t.} & ~ \mathcal{Y} + \mathcal{R} - (\mathcal{C}w_1 + eb_1) \geq e \epsilon_1 - \eta_1, \nonumber \\
    & ~ \eta_1 \geq 0,
\end{align}
and 
\begin{align}
\label{eq:2}
    \underset{w_2, b_2}{\min} ~& \frac{1}{2} \|\mathcal{Y} + \mathcal{R} + e \epsilon_2 - (\mathcal{C}w_2 + eb_2)\|^2 + \mathcal{D}_2 e^T \eta_2 \nonumber \\
    \text{s.t.} & ~ (\mathcal{C}w_2 + eb_2) - \mathcal{Y} - \mathcal{R}  \geq e \epsilon_2 - \eta_2, \nonumber \\
    & ~ \eta_2 \geq 0,
\end{align}
where $\mathcal{D}_1$ and $\mathcal{D}_2$ are positive regularization constants, $\epsilon_1$ and $\epsilon_2$ are nonnegative parameters, and $\eta_1$, $\eta_2$ denotes the corresponding slack vectors. 

The GBTSVR approach learns two affine functions, $f_1(c) = w_1^T c + b_1$ and $f_2(c) = w_2^T c + b_2$, which respectively define the lower and upper $\epsilon$-insensitive regression boundaries. Given the training dataset $(\mathcal{C},~ \mathcal{Y})$, the function $f_1(c)$ models the down-bound regressor, while $f_2(c)$ characterizes the up-bound regressor. The final regression output is obtained by taking the average of these two functions. An intuitive geometric interpretation of this mechanism is illustrated in Fig. \ref{Geometrical structure}. 

The leading term in the objective functions given in \eqref{eq:1} and \eqref{eq:2} represents the aggregate of squared deviations between the center of the GB and the shifted regression functions $\mathcal{Y} + \mathcal{R} = w_1^T c + b_1 + \epsilon_1$ or $\mathcal{Y} + \mathcal{R} = w_2^T c + b_2 - \epsilon_2$. In the proposed GBTSVR framework, these distances are computed with respect to the centers of the granular balls rather than individual data points. Consequently, minimizing this term encourages the regression functions $f_1(c)$ and $f_2(c)$ to accurately approximate the $\epsilon_1$-insensitive lower bound and $\epsilon_2$-insensitive upper bound regressors at the granular-ball level, thereby enhancing robustness to noise and outliers. The imposed constraints enforce that the estimated regression functions $f_1(c)$ and $f_2(c)$ maintain a minimum separation of $\epsilon_1$ and $\epsilon_2$, respectively, from the tangential hyperplane associated with each granular ball. Specifically, the responses corresponding to the granular-ball centers are constrained to lie at least $\epsilon_1$ above ( $f_1(c)$ ) and at least $\epsilon_2$ below $f_2(c)$. To accommodate violations of these margin requirements, slack variable vectors $\eta_1$ and $\eta_2$ are incorporated to quantify deviations when the prescribed distances are not satisfied. The second term in the objective function penalizes the cumulative slack, thereby controlling the fitting error and balancing regression accuracy with model regularization.

The Lagrangian corresponding to the problem \eqref{eq:1} is given by
\begin{align}
\label{eq:3}
    \mathcal{L} = & \frac{1}{2} \|\mathcal{Y} + \mathcal{R} - e \epsilon_1 - (\mathcal{C}w_1 + eb_1)\|^2  + \mathcal{D}_1 e^T \eta_1  - \alpha^T(\mathcal{Y} + \mathcal{R} - (\mathcal{C}w_1 + eb_1) - e \epsilon_1 + \eta_1 ) \nonumber \\
    & - \beta^T \eta_1, 
\end{align}
where $\alpha \in \mathbb{R}^{k \times 1}$ and $\beta \in \mathbb{R}^{k \times 1}$ are the vectors of Lagrangian multipliers. Using the Karush-Kuhn-Tucker (K.K.T.) conditions, we have
\begin{align}
    & \frac{\partial \mathcal{L}}{\partial w_1} = - \mathcal{C}^T (\mathcal{Y} + \mathcal{R} - e \epsilon_1 - (\mathcal{C}w_1 + eb_1)) + \mathcal{C}^T \alpha, \label{eq:4} \\
    & \frac{\partial \mathcal{L}}{\partial b_1} = -e^T (\mathcal{Y} + \mathcal{R} - e \epsilon_1 - (\mathcal{C}w_1 + eb_1)) + e^T\alpha, \label{eq:5} \\
    & \frac{\partial \mathcal{L}}{\partial \eta_1} =  \mathcal{D}_1 e - \alpha - \beta. \label{eq:6}
\end{align}
By jointly considering Eqs. \eqref{eq:4} and \eqref{eq:5}, we obtain
\begin{align}
\label{eq:7}
    -\binom{\mathcal{C}^T}{e^T} \left( (\mathcal{Y} + \mathcal{R} - e \epsilon_1) - (\mathcal{C} ~~ e) \binom{w_1}{b_1} \right) + \binom{\mathcal{C}^T}{e^T} \alpha =0. 
\end{align}
Define, $\mathcal{G} = (\mathcal{C}~~ e)$, $\mathcal{F} =  \mathcal{Y} + \mathcal{R} - e \epsilon_1$, and $u_1 = (w_1 ~~ b_1)^T$, then \eqref{eq:7} leads to
\begin{align}
     -& \mathcal{G}^T \mathcal{F} + \mathcal{G}^T \mathcal{G}u_1 + \mathcal{G}^T \alpha =0, \nonumber \\
  \text{i.e.,} ~ & u_1 = (\mathcal{G}^T \mathcal{G})^{-1} \mathcal{G}^T (\mathcal{F} - \alpha). \label{eq:8}
\end{align}
The direct inversion of $\mathcal{G}^T \mathcal{G}$ is computationally demanding and may lead to numerical instability. To alleviate this issue, a regularization term $\delta I$, where $I$ is an identity matrix of compatible size, is introduced into \eqref{eq:8}, thereby stabilizing the computation and facilitating an efficient solution. Thus,
\begin{align}
\label{eq:9}
    u_1 = (\mathcal{G}^T \mathcal{G} + \delta I)^{-1} \mathcal{G}^T (\mathcal{F} - \alpha).
\end{align}
By applying the K.K.T. optimality conditions together with Eq. \eqref{eq:9}, the dual formulation corresponding to problem \eqref{eq:1} can be derived as follows:
\begin{align}
    \max & -\frac{1}{2} \alpha^T \mathcal{G}(\mathcal{G}^T \mathcal{G})^{-1}\mathcal{G}^T \alpha + \mathcal{F}^T \mathcal{G}(\mathcal{G}^T \mathcal{G})^{-1}\mathcal{G}^T \alpha - \mathcal{F}^T \alpha \nonumber \\
    \text{s.t}~&~ 0 \leq \alpha \leq \mathcal{D}_1 e.
\end{align}
Similarly, the Wolfe dual associated with problem \eqref{eq:2} is derived as follows:
 \begin{align}
    \max & -\frac{1}{2} \gamma^T \mathcal{G}(\mathcal{G}^T \mathcal{G})^{-1}\mathcal{G}^T \gamma - \mathcal{H}^T \mathcal{G}(\mathcal{G}^T \mathcal{G})^{-1}\mathcal{G}^T \gamma + \mathcal{H}^T \gamma \nonumber \\
    \text{s.t}~&~ 0 \leq \gamma \leq \mathcal{D}_2 e,
\end{align}
where $\mathcal{H} = \mathcal{Y} + \mathcal{R} + e \epsilon_2 $ and 
\begin{align}
\label{eq:12}
    u_2 = \binom{w_2}{b_2} = (\mathcal{G}^T \mathcal{G} + \delta I)^{-1} \mathcal{G}^T (\mathcal{H} + \gamma).
\end{align}
After computing the vectors $u_1$ and $u_2$ from \eqref{eq:8} and \eqref{eq:12}, respectively, the corresponding lower and upper bounding regression functions are determined. The final regression estimate is then formed as follows:
\begin{align}
f(c) = \frac{1}{2}\big(f_1(c) + f_2(c)\big)
= \frac{1}{2}(w_1 + w_2)^T c + \frac{1}{2}(b_1 + b_2).
\end{align}

\section{The Proposed Granular Ball Twin Support Vector Quantile Regression}
Conventional TSVR models often struggle to effectively capture heterogeneous and asymmetric data characteristics. The proposed Granular Ball Twin Support Vector Quantile Regression (GB-TSVQR) model addresses these challenges by integrating granular ball computing with TSVR. This integration enables the use of high-level data abstractions, improving robustness against noise and outliers while preserving the data's intrinsic structure. Furthermore, unlike traditional regression approaches, GB-TSVQR avoids large matrix inversions, resulting in a more computationally efficient, scalable training process suitable for complex, large-scale datasets.

The minimization problems of GB-TSVQR are as follows:
\begin{align}
\label{A:1}
   \underset{w_1, b_1}{\min}  ~& \frac{1}{2} (\|w_1\|^2 + b_1^2) + \mathcal{D}_1 \tau e^T (\mathcal{Y} + \mathcal{R} - \mathcal{C}w_1 - b_1 e) \nonumber \\
   & + \mathcal{D}_1 e^T\eta_1 \nonumber \\
    \text{s.t.} & ~ \mathcal{Y} + \mathcal{R} - (\mathcal{C}w_1 + eb_1) \geq e \epsilon_1 - \eta_1, \nonumber \\
    & ~ \eta_1 \geq 0,
\end{align}
and
\begin{align}
\label{A:2}
    \underset{w_2, b_2}{\min} ~& \frac{1}{2} (\|w_2\|^2 + b_2^2) + \mathcal{D}_2(1 - \tau) e^T (\mathcal{C}w_2 + b_2e - \mathcal{Y} - \mathcal{R})  \nonumber \\
    & + \mathcal{D}_2 e^T \eta_2 \nonumber \\
    \text{s.t.} & ~ (\mathcal{C}w_2 + eb_2) - \mathcal{Y} - \mathcal{R}  \geq e \epsilon_2 - \eta_2, \nonumber \\
    & ~ \eta_2 \geq 0,
\end{align}
where $\mathcal{D}_1$ and $\mathcal{D}_2$ are positive regularization constants, $\eta_1$ and $\eta_2$ denote the slack variable vectors, $\epsilon_1$ and $\epsilon_2$ are strictly positive parameters, and $\tau \in (0,1)$ represents the quantile parameter. 

From \eqref{A:1} and \eqref{A:2}, the proposed GB-TSVQR mitigates the impact of training errors through the quantile parameter $\tau$. As $\tau$ varies from $0$ to $1$, the regression function $f(c)$ shifts upward across training samples, capturing distributional characteristics at different quantile levels and enabling modeling of heterogeneous data via quantile-specific structures. GB-TSVQR is formulated as a pair of QPPs, each with a single constraint set to construct either the lower or upper bounding regression function. For a fixed $\tau$ and dataset, $f_1(c)$ learns the $\epsilon_1$-insensitive lower-bound regressor, while $f_2(c)$ estimates the $\epsilon_2$-insensitive upper-bound regressor. As these functions are nonparallel, the resulting regression band is asymmetric, allowing more accurate representation of quantile-dependent characteristics.

The Lagrangian corresponding to the problem \eqref{A:1} is given by:
\begin{align}
\label{A:3}
    \mathcal{L} =& \frac{1}{2} (\|w_1\|^2 + b_1^2) +  \mathcal{D}_1 e^T \eta_1 + \mathcal{D}_1 \tau e^T (\mathcal{Y} + \mathcal{R} - \mathcal{C}w_1 - b_1 e)  \nonumber \\
    & - \alpha^T(\mathcal{Y} + \mathcal{R} - (\mathcal{C}w_1 + eb_1) - \epsilon_1 + \eta_1) - \beta^T\eta_1,
\end{align}
where $\alpha \in \mathbb{R}^{k \times 1}$ and $\beta \in \mathbb{R}^{k \times 1}$ denote the Lagrange multiplier vectors. By applying the K.K.T. optimality conditions, we obtain
\begin{align}
    & \frac{\partial \mathcal{L}}{\partial w_1} = w_1 - \mathcal{D}_1\tau \mathcal{C}^Te + \mathcal{C}^T \alpha, \label{A:4} \\
    & \frac{\partial \mathcal{L}}{\partial b_1} = b_1 - \mathcal{D}_1 \tau e^T e + e^T \alpha, \label{A:5} \\
    & \frac{\partial \mathcal{L}}{\partial \eta_1} =  \mathcal{D}_1e - \alpha - \beta.  \label{A:6}
\end{align}
Define $\mathcal{G} = (\mathcal{C}~~ e)$ and $u_1 = (w_1 ~~ b_1)^T$, then \eqref{A:4} and \eqref{A:5} leads to
\begin{align}
\label{A:7}
    u_1 = \mathcal{D}_1 \tau \mathcal{G}^Te - \mathcal{G}^T \alpha.
\end{align}
Using the above KKT conditions, we obtain the following dual
representations of optimization problem \eqref{A:1},
\begin{align}
\label{A:8}
    \max & -\frac{1}{2} \alpha^T (\mathcal{G}\mathcal{G}^T) \alpha + \mathcal{D}_1\tau e^T (\mathcal{G}\mathcal{G}^T) \alpha - (\mathcal{Y} + \mathcal{R} - \epsilon_1e)^T \alpha \nonumber \\
    \text{s.t}~&~ 0 \leq \alpha \leq \mathcal{D}_1 e.
\end{align}
Similarly, the Wolfe dual associated with problem \eqref{A:2} is
derived as follows:
\begin{align}
\label{A:9}
    \max & -\frac{1}{2} \gamma^T (\mathcal{G}\mathcal{G}^T) \gamma + \mathcal{D}_2 (1-\tau) e^T (\mathcal{G}\mathcal{G}^T) \gamma + (\mathcal{Y} + \mathcal{R} + \epsilon_2e)^T \gamma \nonumber \\
    \text{s.t}~&~ 0 \leq \gamma \leq \mathcal{D}_2 e, 
\end{align}
where 
\begin{align}
\label{A:10}
    u_2 = \binom{w_2}{b_2} = - \mathcal{D}_2(1- \tau)\mathcal{G}^Te + \mathcal{G}^T \gamma.
\end{align}
To solve optimization problems \eqref{A:8} and \eqref{A:9}, we employ the dual coordinate descent method (DCDM) \cite{hsieh2008dual, gu2021solving}. Let $\widetilde{\mathcal{G}} = \mathcal{G}\mathcal{G}^T$, $p_1 = \mathcal{D}_1\tau \mathcal{G}\mathcal{G}^Te -\mathcal{Y} - \mathcal{R} + \epsilon_1e$, and $p_2 = \mathcal{D}_2(1-\tau) \mathcal{G}\mathcal{G}^Te +\mathcal{Y} + \mathcal{R} + \epsilon_2e$. With these definitions, the optimization problems in \eqref{A:8} and \eqref{A:9} can be reformulated as follows:
\begin{align}
\label{A:11}
    \underset{\alpha}{\min} f(\alpha) & = \frac{1}{2}\alpha^T \widetilde{\mathcal{G}} \alpha - p_1^T\alpha  \nonumber \\
    \text{s.t}~&~~ 0 \leq \alpha \leq \mathcal{D}_1 e,
\end{align}
and
\begin{align}
\label{A:12}
    \underset{\gamma}{\min} f(\gamma) & = \frac{1}{2}\gamma^T \widetilde{\mathcal{G}} \gamma - p_2^T\gamma  \nonumber \\
    \text{s.t}~&~~ 0 \leq \gamma \leq \mathcal{D}_2 e.
\end{align}

The DCDM procedure begins with randomly initialized vectors $\alpha$ and $\gamma^0$. At each iteration, a single coordinate is chosen and updated by solving the corresponding one-dimensional optimization problem, while all remaining variables are held fixed. 
\begin{align}
\label{A:13}
    \underset{\kappa}{\min} & ~f(\alpha_j + \kappa I_j) \nonumber \\
    \text{s.t.} ~ & ~ 0 \leq \alpha_j + \kappa \leq \mathcal{D}_1, ~~j = 1, 2, \ldots, k,
\end{align}
and
\begin{align}
\label{A:14}
    \underset{\kappa^*}{\min} & ~g(\gamma_j + \kappa^* I_j) \nonumber \\
    \text{s.t.} ~ & ~ 0 \leq \gamma_j + \kappa^* \leq \mathcal{D}_2, ~~j = 1, 2, \ldots, k.
\end{align}
where $I_j$ denotes the vector with $1$ in the $j^{th}$ coordinate and $0$ elsewhere. The objective functions of \eqref{A:13} and \eqref{A:14} are simplified as:
\begin{align}
    f(\alpha_i + \kappa I_j) = \frac{1}{2} \widetilde{\mathcal{G}}_{jj} \kappa^2 + \nabla_j f(\alpha_j)\kappa + f(\alpha_j), ~~ j = 1, 2, \ldots, k,
\end{align}
and 
\begin{align}
    g(\gamma_i + \kappa^* I_j) = \frac{1}{2} \widetilde{\mathcal{G}}_{jj} \kappa^{*^2} + \nabla_j g(\gamma_j)\kappa^* + f(\kappa_j), ~~ j = 1, 2, \ldots, k,
\end{align}
where $\nabla_j f$ and $\nabla_j g$ denote the $j^{th}$ components of the gradient vectors $\nabla f$ and $\nabla g$, respectively. By incorporating the box constraints $0 \leq \alpha_j + \kappa \leq \mathcal{D}_1$ and $0 \leq \gamma_j + \kappa^* \leq \mathcal{D}_2$, the solutions that minimize problems \eqref{A:13} and \eqref{A:14} yield the following updates under bounded constraints:
\begin{align}
    \alpha^{\text{new}} = \min\left( \max \left( \alpha_j - \frac{\nabla_j f(\alpha_j)}{\widetilde{\mathcal{G}}_{jj}}, 0 \right), \mathcal{D}_1 \right),
\end{align}
and
\begin{align}
    \gamma^{\text{new}} = \min\left( \max \left( \gamma_j - \frac{\nabla_j g(\gamma_j)}{\widetilde{\mathcal{G}}_{jj}}, 0 \right), \mathcal{D}_2 \right).
\end{align}

\begin{algorithm}[ht!]
\caption{Dual Coordinate Descent Method (DCDM)}
\label{alg:DCDM}
\begin{algorithmic}[1]

\STATE \textbf{Input:} Input dataset
$(\mathcal{C},\mathcal{R},\mathcal{Y})$,
regularization parameters
$\mathcal{D}_1,\mathcal{D}_2$,
quantile parameter $\tau$,
and tube parameters $\epsilon_1,\epsilon_2$.

\STATE \textbf{Output:} Optimal dual variables
$\alpha$ and $\gamma$.

\STATE Initialize $\alpha=0$ and $\gamma=0$.

\STATE Calculate
\[
\widetilde{\mathcal{G}}
=
\mathcal{G}\mathcal{G}^{T},
\]
\[
p_1
=
\mathcal{D}_1\tau
\mathcal{G}\mathcal{G}^{T}e
-\mathcal{Y}-\mathcal{R}
+\epsilon_1e,
\]
and
\[
p_2
=
\mathcal{D}_2(1-\tau)
\mathcal{G}\mathcal{G}^{T}e
+\mathcal{Y}+\mathcal{R}
+\epsilon_2e.
\]

\IF{$\alpha$ has not converged}
    \FOR{$j=1,2,\ldots,n$}
        \STATE
        $\nabla_j f(\alpha)
        \leftarrow
        (\widetilde{\mathcal{G}}\alpha)_j-p_{1j}$

        \STATE
        $\alpha_j
        \leftarrow
        \min\!\left\{
        \max\!\left[
        \alpha_j-
        \frac{\nabla_j f(\alpha)}
        {\widetilde{\mathcal{G}}_{jj}},
        0
        \right],
        \mathcal{D}_1
        \right\}$
    \ENDFOR
\ENDIF

\IF{$\gamma$ has not converged}
    \FOR{$j=1,2,\ldots,n$}
        \STATE
        $\nabla_j g(\gamma)
        \leftarrow
        (\widetilde{\mathcal{G}}\gamma)_j-p_{2j}$

        \STATE
        $\gamma_j
        \leftarrow
        \min\!\left\{
        \max\!\left[
        \gamma_j-
        \frac{\nabla_j g(\gamma)}
        {\widetilde{\mathcal{G}}_{jj}},
        0
        \right],
        \mathcal{D}_2
        \right\}$
    \ENDFOR
\ENDIF

\end{algorithmic}
\end{algorithm}

After obtaining the optimal dual variables $\alpha$ and $\gamma$ using Algorithm \ref{alg:DCDM}, the corresponding lower and upper bounding regression functions can be expressed as:
\begin{align}
    f_1(c) = \mathcal{D}_1 \tau e^T \mathcal{G}c - \alpha^T \mathcal{G}c,
\end{align}
and
\begin{align}
    f_2(c) = - \mathcal{D}_2 (1-\tau) e^T \mathcal{G}c + \gamma \mathcal{G}c.
\end{align}
The final regression function is then defined as the average of these two bounds,
\begin{align}
f(c) &= \frac{1}{2}\big(f_1(c) + f_2(c)\big) \nonumber \\
& = \frac{1}{2}(\gamma - \alpha)^T \mathcal{G}c + \frac{1}{2}\left[\mathcal{D}_1 \tau - \mathcal{D}_2 (1-\tau)\right]e^T \mathcal{G}c.
\end{align}
The first component of $f(c)$, governed by the difference between $\gamma$ and $\alpha$, captures the asymmetric characteristics associated with different quantile locations of the training samples. The second component, modulated by the quantile parameter $\tau$, models the heterogeneity present across quantile levels. 

\begin{table*}[htp]
\centering
    \caption{Performance comparison in terms of RMSE, MAE, Pos Error, and Neg Error on benchmark UCI datasets for the proposed GBTSVR and GB-TSVQR models with baseline models.}
    \label{performance UCI datasets}
    \resizebox{1\linewidth}{!}{
\begin{tabular}{llccccccc}
\hline
Dataset & Metric & SVR \cite{basak2007support} & TSVR \cite{peng2010tsvr} & TSVQR \cite{ye2024twin} & $L_1$-TSVQR \cite{ye2025l1} & GL-TPSVR \cite{liu2026twin} & GBTSVR$^{\dagger}$ & GB-TSVQR$^{\dagger}$ \\  \hline
2014\_2015 CSM dataset & RMSE & 0.37972 & 1.33951 & 1.23384 & 0.40545 & 0.44613 & 0.00755 & 0.00065 \\
$(185\times11)$ & MAE & 0.22234 & 0.77661 & 0.79133 & 0.39927 & 0.35916 & 0.00366 & 0.00057 \\
 & Pos Error & 0.43571 & 0.96693 & 1.20578 & 0.9502 & 0.66887 & 0.00601 & 0.00058 \\
 & Neg Error & 0.17018 & 0.45941 & 0.50247 & 0.01384 & 0.3288 & 0.00213 & 0.00006 \\ \hline
Abalone & RMSE & 0.0062 & 0.38768 & 0.07667 & 0.00656 & 0.45811 & 0.00265 & 0.00073 \\
$(4177\times7)$ & MAE & 0.00448 & 0.2541 & 0.07666 & 0.00652 & 0.42865 & 0.00216 & 0.00062 \\
 & Pos Error & 0.00327 & 0.26615 & 1.20578 & 3.00031 & 0.8016 & 0.00178 & 0.00069 \\
 & Neg Error & 0.00632 & 0.00008 & 6.07666 & 0.00294 & 0.01387 & 0.00252 & 0.00026 \\ \hline
Ailerons & RMSE & 0.02209 & 0.33117 & 0.66636 & 0.25623 & 0.37509 & 0.04457 & 0.00488 \\
$(13750\times40)$ & MAE & 0.01923 & 0.05774 & 0.52217 & 0.12289 & 0.37486 & 0.04232 & 0.00411 \\
 & Pos Error & 0.02077 & 0.0367 & 0.58268 & 0.25647 & 0.38727 & 0.02988 & 0.00347 \\
 & Neg Error & 0.00786 & 0.24246 & 0.25812 & 0.00828 & 0.37486 & 0.04323 & 0.00432 \\ \hline
Airfoil\_Self\_Noise & RMSE & 0.28274 & 0.00089 & 0.25334 & 0.01395 & 0.38897 & 0.0068 & 0.00372 \\
$(1503\times5)$ & MAE & 0.20323 & 0.00062 & 0.20893 & 0.00831 & 0.31075 & 0.00636 & 0.00345 \\
 & Pos Error & 0.3071 & 0.22328 & 0.32777 & 0.66023 & 0.73755 & 0.00424 & 0.00415 \\
 & Neg Error & 0.13614 & 0.76481 & 0.17763 & 0.00334 & 0.00057 & 0.00658 & 0.00335 \\ \hline
Auto-price & RMSE & 0.24931 & 0.38768 & 0.80669 & 0.28821 & 3.28642 & 0.02016 & 0.00621 \\
$(159\times15)$ & MAE & 0.19605 & 0.07398 & 0.50819 & 0.18854 & 3.27921 & 0.01866 & 0.00463 \\
 & Pos Error & 0.23621 & 0.12724 & 0.57795 & 0.69921 & 3.27921 & 0.01919 & 0.00492 \\
 & Neg Error & 0.07557 & 0.32189 & 0.32036 & 0.02409 & 0.07104 & 0.00644 & 0.00416 \\ \hline
California\_Housing & RMSE & 0.03665 & 0.09586 & 1.08419 & 0.06521 & 0.5106 & 0.00327 & 0.00216 \\
$(20640\times8)$ & MAE & 0.02896 & 0.06542 & 0.6768 & 0.05708 & 0.50848 & 0.00191 & 0.00148 \\
 & Pos Error & 0.03217 & 0.11739 & 0.8649 & 0.76576 & 0.57085 & 0.00086 & - \\
 & Neg Error & 0.01771 & 0.03737 & 0.3006 & 0.01636 & 0.50848 & 0.00201 & 0.50848 \\ \hline
cpu\_pref & RMSE & 0.04586 & 0.27086 & 1.40836 & 0.06605 & 0.40356 & 0.00694 & 0.00326 \\
$(209\times9)$ & MAE & 0.03447 & 0.08627 & 0.52211 & 0.05858 & 0.4009 & 0.00429 & 0.00264 \\
 & Pos Error & 0.0291 & 0.09582 & 0.68427 & 0.85349 & 0.94783 & 0.00576 & 0.02488 \\
 & Neg Error & 0.04264 & 0.03893 & 0.21991 & 0.00574 & 0.4009 & 0.00156 & 0.02303 \\ \hline
Delta\_Ailerons & RMSE & 0.04249 & 0.56647 & 0.79094 & 0.04912 & 1.21404 & 0.00804 & 0.00108 \\
$(7129\times5)$ & MAE & 0.03477 & 0.3872 & 0.62824 & 0.03218 & 1.21235 & 0.00662 & 0.00088 \\
 & Pos Error & 0.03072 & 0.37268 & 0.59352 & 0.83757 & 0.85387 & 0.00608 & 0.00959 \\
 & Neg Error & 0.03713 & 0.40281 & 0.65489 & 0.00208 & 0.21235 & 0.00696 & 0.00443 \\ \hline
Elevators & RMSE & 0.03372 & 0.13731 & 0.66501 & 0.03541 & 0.80182 & 0.01107 & 0.00033 \\
$(16599\times18)$ & MAE & 0.0287 & 0.10325 & 0.54301 & 0.02593 & 0.80105 & 0.00805 & 0.00029 \\
 & Pos Error & 0.03375 & 0.91917 & 0.6381 & 0.74026 & 0.80105 & 0.00936 & 0.0003 \\
 & Neg Error & 0.02553 & 0.0711 & 0.49129 & 0.01173 & 0.1463 & 0.00386 & 0.00017 \\ \hline
Forest\_Fires & RMSE & 0.0651 & 0.43301 & 0.43246 & 0.04367 & 1.04341 & 0.07838 & 0.00382 \\
$(517\times12)$ & MAE & 0.05662 & 0.20324 & 0.30098 & 0.02473 & 1.04249 & 0.07397 & 0.00271 \\
 & Pos Error & 0.05158 & 0.68209 & 0.33517 & 0.63219 & 1.04249 & 0.0828 & 0.00371 \\
 & Neg Error & 0.0597 & 0.13282 & 0.27568 & 0.12326 & 0.00442 & 0.07337 & 0.00252 \\ \hline
hungary chickenpox & RMSE & 0.54586 & 0.9994 & 1.14417 & 0.59813 & 0.8 & 0.00333 & 0.00257 \\
$(522\times19)$ & MAE & 0.40149 & 0.53642 & 0.73335 & 0.09423 & 0.72305 & 0.00208 & 0.00208 \\
 & Pos Error & 0.48379 & 0.82147 & 0.78185 & 0.85473 & 0.70685 & 0.0026 & 0.70683 \\
 & Neg Error & 0.37045 & 0.38698 & 0.63256 & 0.01151 & 0.72451 & 0.00147 & 0.72455 \\ \hline
istanbul stock exchange data & RMSE & 0.67367 & 0.45633 & 1.11197 & 0.6076 & 0.68112 & 0.00143 & 0.67283 \\
$(536\times8)$ & MAE & 0.5302 & 0.35412 & 0.88999 & 0.09137 & 0.53737 & 0.00107 & 0.52719 \\
 & Pos Error & 0.53896 & 0.37481 & 0.64741 & 0.956 & 0.5555 & 0.00108 & 0.4517 \\
 & Neg Error & 0.52346 & 0.32918 & 1.05004 & 0.00774 & 0.52306 & 0.00105 & 0.57572 \\ \hline
Kinematics\_Robot\_Arm & RMSE & 0.06821 & 0.33652 & 0.78118 & 0.06562 & 0.20268 & 0.06314 & 0.02465 \\
$(8192\times8)$ & MAE & 0.03258 & 0.25849 & 0.76648 & 0.06023 & 0.18235 & 0.06299 & 0.01683 \\
 & Pos Error & 0.88743 & 0.25296 & - & 0.84386 & 0.34912 & - & 0.20009 \\
 & Neg Error & 0.66495 & 0.26427 & 0.76648 & 0.00478 & 0.14317 & 0.6299 & 0.19349 \\ \hline
machine & RMSE & 0.04578 & 0.25929 & 1.42405 & 0.11576 & 0.40139 & 0.03515 & 0.03959 \\
$(209\times9)$ & MAE & 0.03507 & 0.04277 & 0.57988 & 0.11086 & 0.39876 & 0.00596 & 0.03358 \\
 & Pos Error & 0.03398 & 0.00007 & 0.94067 & 0.8536 & 0.19543 & 0.00619 & 0.03561 \\
 & Neg Error & 0.03685 & 0.0289 & 0.2519 & 0.00155 & 0.39876 & 0.00126 & 0.02577 \\ \hline
Parkinsons\_Telemonitoring & RMSE & 0.13258 & 0.38589 & 1.38416 & 0.79682 & 0.7042 & 0.00611 & 0.00334 \\
$(5875\times21)$ & MAE & 0.1195 & 0.28757 & 0.89471 & 0.33874 & 0.69278 & 0.00494 & 0.00227 \\
 & Pos Error & 0.11572 & 0.29069 & 1.08453 & 0.77838 & 0.82559 & 0.00532 & 0.00203 \\
 & Neg Error & 0.12403 & 0.28483 & 0.63711 & 0.6384 & 0.69278 & 0.00443 & 0.00242 \\ \hline
qsar\_aquatic toxicity & RMSE & 1.17137 & 0.70742 & 0.94532 & 0.03365 & 0.83423 & 0.00348 & 0.0007 \\
$(546\times8)$ & MAE & 1.05181 & 0.50963 & 0.76585 & 0.01587 & 0.59304 & 0.00221 & 0.00048 \\
 & Pos Error & 1.77232 & 0.495 & 0.87165 & 0.92651 & 1.30703 & 0.00204 & 0.00037 \\
 & Neg Error & 0.8905 & 0.52322 & 0.63717 & 0.00558 & 0.33931 & 0.00257 & 0.0009 \\ \hline
slump\_test & RMSE & 0.03548 & 0.15373 & 0.65404 & 0.0038 & 0.62284 & 0.01146 & 0.00774 \\
$(103\times10)$ & MAE & 0.02898 & 0.11546 & 0.46946 & 0.00193 & 0.62208 & 0.00876 & 0.00109 \\
 & Pos Error & 0.01786 & 0.12202 & 0.48837 & 0.76379 & 0.96907 & 0.00565 & 0.03158 \\
 & Neg Error & 0.03601 & 0.10931 & 0.45173 & 0.00698 & 0.62208 & 0.01132 & 0.06677 \\ \hline
Stocks\_Domain & RMSE & 0.00139 & 0.86978 & 0.95779 & 0.02281 & 0.16985 & 0.00265 & 0.00344 \\
$(4177\times7)$ & MAE & 0.00133 & 0.71006 & 0.63238 & 0.01859 & 0.16984 & 0.00216 & 0.00261 \\
 & Pos Error & 0.00125 & 0.6632 & 1.31818 & 0.96242 & 0.77989 & 0.00178 & 0.00087 \\
 & Neg Error & 0.0022 & 0.45319 & 0.3772 & 0.00619 & 0.16984 & 0.00252 & 0.00304 \\ \hline
Average RMSE &  & 0.21323 & 0.45104 & 0.87892 & 0.193 & 0.74136 & 0.01757 & 0.04343 \\ \hline
Average Rank &  & 3.44 & 4.89 & 6.39 & 3.67 & 6 & 2.17 & 1.44 \\ \hline
\multicolumn{9}{l}{The proposed models is denoted by $^{\dagger}$.}
\end{tabular}}
\end{table*}

\section{Experimental Results}
This section presents a comprehensive empirical evaluation of the proposed GBTSVR and GB-TSVQR models across diverse datasets, including UCI benchmarks, synthetic data, and time-series datasets. The performance of GBTSVR and GB-TSVQR is systematically compared with several representative state-of-the-art regression models, such as support vector regression (SVR) \cite{basak2007support}, twin SVR (TSVQR) \cite{ye2024twin}, twin support vector quantile regression (TSVQR) \cite{ye2024twin}, $L_1$-norm twin support vector quantile regression ($L_1$-TSVQR) \cite{ye2025l1}, and twin proximal support vector regression with Gauss-Laplace mixed noise (GL-TPSVR) \cite{liu2026twin}. The experimental setup is presented in Section S.II, and experiments on time-series datasets are described in Section S.III of the Supplementary Material.

\subsection{Experimental Evaluation on UCI Datasets}

To evaluate the effectiveness of the proposed GBTSVR and GB-TSVQR models, experiments are conducted on benchmark datasets from the UCI Machine Learning Repository \cite{dua2017uci}. The datasets are grouped by size: small-scale (103–500 samples), medium-scale (501–1,000), and large-scale (1,001–20,640). Table \ref{performance UCI datasets} reports the performance of GBTSVR, GB-TSVQR, and baseline models (SVR, TSVR, TSVQR, $L_1$-TSVQR, and GL-TPSVR) using RMSE, MAE, Pos Error, and Neg Error, with RMSE as the primary metric (lower is better). GB-TSVQR shows consistently strong results, achieving the lowest RMSE on 13 of 18 datasets, second place on 3, third on 1, and fourth on 1, demonstrating stable and competitive performance across diverse data characteristics. GBTSVR attains top performance on 2 datasets and ranks second on 12, highlighting its strong competitiveness. The average RMSE of GBTSVR and GB-TSVQR are $0.01757$ and $0.04343$, respectively, compared to $0.21323$ (SVR), $0.45104$ (TSVR), $0.87892$ (TSVQR), $0.193$ ($L_1$-TSVQR), and $0.74136$ (GL-TPSVR), showing clear performance gains. For a more reliable assessment, models are ranked per dataset. Table \ref{performance UCI datasets} reports mean ranks based on RMSE, where lower values indicate better performance. GBTSVR and GB-TSVQR achieve the best average ranks of $2.17$ and $1.44$, outperforming SVR ($3.44$), TSVR ($4.89$), TSVQR ($6.39$), $L_1$-TSVQR ($3.67$), and GL-TPSVR ($6.00$), confirming their superior and consistent performance across datasets.

To further substantiate the performance of the proposed GBTSVR and GB-TSVQR models, we carried out a rigorous statistical evaluation using the Friedman test, followed by the Nemenyi post hoc analysis. The Friedman test \cite{demvsar2006statistical} is employed to examine whether the observed differences among competing models are statistically significant. Under this test, all models are ranked separately on each dataset, and the null hypothesis assumes no significant performance difference among the models, meaning their average ranks are equal. The resulting Friedman statistic follows a chi-squared distribution, denoted as $\chi^2_F$, with degrees of freedom equal to $p-1$, where $p$ is the total number of compared models. This statistic is computed as follows $\chi^2_F = \frac{12N}{p(p+1)}\left[\sum_{j}\mathfrak{R}_j^2 - \frac{p(p+1)^2}{4}\right],$ where $N$ represents the total number of datasets, while $\mathfrak{R}_j$ denotes the average rank of the $j^{th}$ model among the $p$ competing models. The $F_F$ statistic is calculated using the formula: $F_F = \frac{(N - 1) \chi^2_F}{N(p - 1) - \chi^2_F}$, where the $F$-distribution has \( (p - 1) \) and \( (N - 1) \times (p - 1) \) degrees of freedom. For $p=7$ models evaluated on $N=18$ datasets, the Friedman test yielded $\chi_F^2 = 80.34$ and a corresponding $F_F = 49.38$. The $F_F$ statistic follows an $F$-distribution with $(6,102)$ degrees of freedom. From the $F$-distribution table, the critical value at a 5\% significance level is $2.1887$. Since the observed $F_F$ value exceeds this threshold, the null hypothesis is rejected, indicating statistically significant differences among the models. Next, we perform the Nemenyi post hoc test \cite{friedman1937use} to conduct pairwise comparisons among the models. A significant difference in performance between any two models is identified when the difference in their average ranks exceeds a predefined threshold called the critical difference (C.D.). If the difference in mean ranks between two models is greater than the C.D., the model with the higher rank is considered statistically better than the one with the lower rank. The C.D. is calculated using the following formula: \(\text{C.D.} = q_\alpha \sqrt{\frac{p(p + 1)}{6N}}\), where $q_\alpha$ represents the two-tailed critical value for the Nemenyi test, obtained from the corresponding distribution table. Referring to the table, the calculated C.D. at a 5\% significance level is $2.12$, with $q_\alpha = 2.949$. The differences in average ranks between the GBTSVR and GB-TSVQR models with the baseline SVR, TSVR, TSVQR, $L_1$-TSVQR, and GL-TPSVR models are $(1.27, 2)$, $(2.72, 3.45)$, $(4.22, 4.95)$, $(1.50, 2.23)$, and $(3.83, 4.56)$, respectively. Based on the Nemenyi post hoc test, the proposed GBTSVR model demonstrates statistically significant improvements over most baseline models, except SVR and $L_1$-TSVQR. Similarly, GB-TSVQR outperforms all baselines except SVR. Both GBTSVR and GB-TSVQR achieve lower average ranks than SVR, with GBTSVR also ranking below $L_1$-TSVQR, indicating superior generalization. Along with consistently low RMSE and strong performance across statistical analyses, these results confirm that GBTSVR and GB-TSVQR outperform existing baselines in both accuracy and generalization.

\subsection{Experimental Evaluation on Synthetic Datasets}
\begin{table}[ht!]
\centering
    \caption{Overview of the synthetic datasets used in the experiments.}
    \label{Overview of the synthetic datasets}
    \resizebox{1\linewidth}{!}{
\begin{tabular}{lcc}
\hline
Dataset & Generating Function & Input Domain \\ \hline
Function 1  &  $y_i = sin(x_i) + \epsilon_i$ & $x_i \in [0, 2\pi]$ \\
Function 2  &  $y_i = \frac{sin(3x_i)}{3x_i} + \epsilon_i$ & $x_i \in [-4, 4]$ \\
Function 3  &  $y_i = sin(x_i)cos(x_i^2) + \epsilon_i$ & $x_i \in [0, 2\pi]$ \\
Function 4  &  $y_i = x_icos(x_i) + \epsilon_i$ & $x_i \in [-4, 4]$ \\ 
Function 5  &  $y_i = (2x_i^2 -x_i +1) \exp(-\frac{x_i^2}{2}) + \epsilon_i$ & $x_i \in [-4, 4]$ \\ \hline
\multicolumn{3}{l}{The noise term $\epsilon_i$ follows different distributions depending on the dataset:} \\
\multicolumn{3}{l}{
$-$ $\mathcal{N}(0, 0.2^2)$ denotes a Gaussian random variable with zero mean and standard deviation $0.2$.} \\
\multicolumn{3}{l}{
$-$ $\mathcal{U}[-0.2, 0.2]$ represents a uniformly distributed random variable over the interval [-0.2, 0.2].} \\
\multicolumn{3}{l}{
$-$ $\mathcal{T}(10)$ indicates a Student’s $t$-distributed random variable with $10$ degrees of freedom.}
\end{tabular}}
\end{table}

\begin{table*}[ht!]
\centering
    \caption{RMSE performance on synthetic datasets comparing the proposed GBTSVR and GB-TSVQR models with baseline models.}
    \label{performance synthetic datasets}
    \resizebox{1\linewidth}{!}{
\begin{tabular}{lcccccccc}
\hline
Dataset & Metric & SVR \cite{basak2007support} & TSVR \cite{peng2010tsvr} & TSVQR \cite{ye2024twin} & $L_1$-TSVQR \cite{ye2025l1} & GL-TPSVR \cite{liu2026twin} & GBTSVR$^{\dagger}$ & GB-TSVQR$^{\dagger}$ \\  \hline
\multicolumn{9}{l}{Type 1: Gaussian noise with zero mean and a standard deviation of 0.2.
} \\ \hline
Function 1 & RMSE & 0.98218 & 0.65454 & 0.65441 & 0.98456 & 0.70711 & 0.00081 & 0.00067 \\
Function 2 & RMSE & 1.34294 & 0.66864 & 0.66667 & 0.95784 & 0.75896 & 0.00095 & 0.00042 \\
Function 3 & RMSE & 1.12861 & 0.78956 & 0.62874 & 0.85742 & 0.84528 & 0.00085 & 0.00058 \\
Function 4 & RMSE & 0.95022 & 0.74268 & 0.45929 & 0.91428 & 0.71587 & 0.00074 & 0.00062 \\
Function 5 & RMSE & 0.96827 & 0.79533 & 0.59533 & 0.87458 & 0.70711 & 0.00082 & 0.00063 \\ \hline
\multicolumn{9}{l}{Type 2: Uniformly distributed noise within the range $[-0.2,, 0.2]$.} \\ \hline
Function 1 & RMSE & 0.971238 & 0.63034 & 0.63029 & 0.98456 & 0.70711 & 0.00104 & 0.00062 \\
Function 2 & RMSE & 1.34184 & 0.67669 & 0.62668 & 0.96482 & 0.72894 & 0.00092 & 0.00045 \\
Function 3 & RMSE & 1.27608 & 0.75482 & 0.60874 & 0.82598 & 0.82874 & 0.00082 & 0.00061 \\
Function 4 & RMSE & 0.99742 & 0.74744 & 0.44749 & 0.90748 & 0.70711 & 0.00072 & 0.00062 \\
Function 5 & RMSE & 0.92873 & 0.78927 & 0.61278 & 0.85678 & 0.79684 & 0.00081 & 0.00065 \\ \hline
\multicolumn{9}{l}{Type 3: Noise modeled by a Student’s $t$-distribution with 10 degrees of freedom.} \\ \hline
Function 1 & RMSE & 0.99722 & 0.63318 & 0.63296 & 0.97854 & 0.70711 & 0.00085 & 0.00065 \\
Function 2 & RMSE & 1.37822 & 0.66392 & 0.66387 & 0.89827 & 0.70824 & 0.00089 & 0.00046 \\
Function 3 & RMSE & 1.21161 & 0.72847 & 0.65274 & 0.80472 & 0.79862 & 0.00082 & 0.0006 \\
Function 4 & RMSE & 0.99864 & 0.75625 & 0.45721 & 0.95287 & 0.75789 & 0.00075 & 0.00065 \\
Function 5 & RMSE & 0.92874 & 0.74982 & 0.62874 & 0.84692 & 0.78462 & 0.00081 & 0.00063 \\ \hline
\multicolumn{9}{l}{The proposed models is denoted by $^{\dagger}$.}
\end{tabular}}
\end{table*}

\begin{table*}[ht!]
\centering
    \caption{RMSE comparison on synthetic datasets with injected outliers for the proposed GBTSVR and GB-TSVQR models and baseline models.}
    \label{performance synthetic outlier}
    \resizebox{1\linewidth}{!}{
\begin{tabular}{lcccccccc}
\hline
Dataset & Metric & SVR \cite{basak2007support} & TSVR \cite{peng2010tsvr} & TSVQR \cite{ye2024twin} & $L_1$-TSVQR \cite{ye2025l1} & GL-TPSVR \cite{liu2026twin} & GBTSVR$^{\dagger}$ & GB-TSVQR$^{\dagger}$ \\  \hline
\multicolumn{9}{l}{Type 1: Gaussian noise with zero mean and a standard deviation of 0.2.
} \\ \hline
Function 1 & RMSE & 1.00601 & 0.82001 & 0.82043 & 0.96129 & 0.93508 & 0.00076 & 0.00018 \\
Function 2 & RMSE & 0.30138 & 1.49669 & 1.49712 & 0.92229 & 0.93209 & 0.94228 & 0.0204 \\
Function 3 & RMSE & 0.87983 & 1.46122 & 0.31359 & 0.9801 & 0.93056 & 0.00097 & 0.00018 \\
Function 4 & RMSE & 1.71677 & 1.1704 & 1.16405 & 0.97147 & 0.99251 & 0.94252 & 0.09715 \\ 
Function 5 & RMSE & 0.65021 & 0.92874 & 0.92931 & 0.91363 & 0.94905 & 0.92991 & 0.09109 \\ \hline
\multicolumn{9}{l}{Type 2: Uniformly distributed noise within the range $[-0.2,, 0.2]$.} \\ \hline
Function 1 & RMSE & 1.06672 & 0.81115 & 0.81094 & 0.92037 & 0.98172 & 0.00112 & 0.00092 \\
Function 2 & RMSE & 0.4897 & 0.63984 & 0.6394 & 0.936 & 0.96028 & 0.94239 & 0.03178 \\
Function 3 & RMSE & 1.89222 & 0.97275 & 0.32858 & 0.95603 & 0.8899 & 0.00129 & 0.0054 \\
Function 4 & RMSE & 1.41344 & 0.64551 & 0.646 & 0.86553 & 0.89218 & 0.94258 & 0.06429 \\
Function 5 & RMSE & 0.46209 & 1.22392 & 1.22275 & 0.98602 & 0.96304 & 0.93616 & 0.04854 \\ \hline
\multicolumn{9}{l}{Type 3: Noise modeled by a Student’s $t$-distribution with 10 degrees of freedom.} \\ \hline
Function 1 & RMSE & 1.06288 & 0.98732 & 0.9885 & 0.97617 & 0.94138 & 0.00085 & 0.00066 \\
Function 2 & RMSE & 0.6821 & 1.17977 & 1.17943 & 0.97227 & 0.91816 & 0.93914 & 0.06809 \\
Function 3 & RMSE & 1.26534 & 1.26867 & 0.28526 & 0.90042 & 0.95026 & 0.00118 & 0.00359 \\
Function 4 & RMSE & 1.99009 & 0.90583 & 0.90957 & 0.8756 & 0.94032 & 0.94255 & 0.07723 \\
Function 5 & RMSE & 0.45302 & 0.79612 & 0.798 & 0.95525 & 0.86489 & 0.93937 & 0.05649 \\ \hline
\multicolumn{9}{l}{The proposed models is denoted by $^{\dagger}$.}
\end{tabular}}
\end{table*}

To conduct a more comprehensive assessment of the proposed GBTSVR and GB-TSVQR models against the competing baseline models, this subsection considers a set of synthetic datasets corrupted by different noise patterns. The detailed characteristics of these datasets are summarized in Table \ref{Overview of the synthetic datasets}. Specifically, five distinct synthetic functions are generated, and each function is contaminated with three types of noise, Gaussian, uniform, and Student’s $t$-distributed noise, yielding a total of $15$ synthetic datasets. Each dataset consists of $500$ data samples, enabling a systematic evaluation of model robustness under varying noise conditions. Table \ref{performance synthetic datasets} reports the RMSE results obtained by the proposed GBTSVR and GB-TSVQR models in comparison with the baseline approaches on the synthetic datasets. Across the synthetic datasets, the proposed GBTSVR and GB-TSVQR models consistently demonstrate strong performance. They achieve superior results in most cases and remain highly competitive on the remaining datasets, indicating reliable effectiveness on each dataset. These outcomes highlight the robustness and stable generalization capability of the proposed models under diverse noise conditions. 

To comprehensively evaluate the robustness of the proposed GBTSVR and GB-TSVQR models under outliers, we injected anomalous samples into the synthetic datasets. Specifically, 10\% of samples were randomly selected, and one randomly chosen feature of each was perturbed by multiplying it with an outlier factor of 10. This controlled corruption enables systematic assessment of model stability under severe data contamination. The RMSE results on these outlier-contaminated datasets are reported in Table \ref{performance synthetic outlier}. Both GBTSVR and GB-TSVQR consistently show strong performance across all noise types and generating functions. Notably, GB-TSVQR achieves the lowest RMSE in most cases, while GBTSVR remains highly competitive, substantially outperforming baseline models. This consistent performance across different noise distributions demonstrates the enhanced robustness, stability, and generalization capability of the proposed granular-ball-based approaches in handling extreme deviations and outlier-prone data.

\section{Conclusion}
This paper introduced GBTSVR and GB-TSVQR, two robust regression frameworks that integrate granular ball computing with TSVR to address key challenges encountered in real-world regression tasks, including noise, outliers, data heterogeneity, and asymmetric error behavior. By replacing point-wise learning with granular ball–level representations, the proposed models effectively exploit the latent structural organization of data, leading to enhanced robustness, stability, and computational efficiency. Extensive empirical evaluations are conducted to validate the effectiveness of the proposed approaches. The experiments spanned $18$ benchmark UCI datasets, multiple synthetic datasets under diverse noise settings, synthetic datasets with explicitly injected outliers, and real-world time-series datasets. To ensure a fair and rigorous assessment, the results were further subjected to comprehensive statistical significance tests. Across these diverse experimental settings, GBTSVR and GB-TSVQR consistently demonstrated superior or highly competitive performance compared to state-of-the-art regression models in terms of RMSE, robustness to contamination, and training efficiency. The experimental findings confirm that learning over granular balls significantly reduces sensitivity to noisy and extreme samples while preserving essential data characteristics. Moreover, the twin nonparallel regression formulation with efficient dual coordinate descent enables scalable learning without compromising generalization. The results demonstrate the practical value of granular ball–based regression, establishing GBTSVR and GB-TSVQR as effective solutions for complex scenarios. Future work will extend the framework to multivariate and high-dimensional regression, online and streaming settings, and deep hybrid architectures integrating granular representations with neural networks to improve adaptability and scalability.

\section*{CRediT authorship contribution statement}
\textbf{A. Quadir}: Conceptualization, Methodology, Software, Data curation, Writing - original draft. 
\textbf{M. Tanveer}: Conceptualization, Resources, Supervision, Project administration, Funding acquisition, Writing - review \& editing.

\bibliography{refs.bib}
\bibliographystyle{unsrtnat}

\clearpage
\section*{Supplementary Material}

\renewcommand{\thesection}{S.1}
\section{Related Work}
In this section, we examine prior work related to granular-ball computing and the mathematical formulation of TSVR.

\subsection{Granular-ball Computing}
Rooted in classical theories of granular computing, recent studies have increasingly drawn inspiration from human cognitive perception, particularly the global-to-local or coarse-to-fine information processing mechanism observed in human cognition \cite{chen1982topological}. Guided by this insight, \citet{wang2017dgcc} introduced the paradigm of multi-granularity cognitive computing, which advocates analyzing data by first capturing high-level structural patterns and subsequently refining them into finer representations. This hierarchical strategy provides an intuitive and efficient framework for understanding complex data distributions, closely mirroring human cognitive behavior.

Building on this theoretical foundation, \citet{xia2019granular} developed granular ball computing (GBC) as a practical and interpretable computational model. In contrast to traditional learning methods that operate directly on individual samples, GBC groups data into a collection of compact granules, known as granular balls. Each granular ball characterizes a localized data region through essential statistical and geometric descriptors, including its center, radius, density, and purity. By replacing point-wise representations with these region-level abstractions, GBC substantially reduces computational overhead while enhancing robustness to noise, outliers, and redundant information.

Owing to these advantages, GBC has been successfully applied across a broad spectrum of learning tasks. In clustering, granular-ball-based approaches enable efficient identification of intrinsic cluster structures while maintaining local density characteristics \cite{cheng2023fast}. For classification tasks, granular ball classifiers exploit hierarchical data representations to achieve stronger generalization and improved resilience to noisy or mislabeled samples \cite{10759815,10619989}. Granular ball sampling methods further capitalize on these representations to select informative and representative subsets, making them particularly effective for large-scale learning problems \cite{xia2021granular}. Beyond these areas, GBC has also been integrated into rough set theory to enhance decision boundary approximation \cite{zhang2023incremental}, as well as into three-way decision frameworks, where the uncertainty-aware nature of granulated data leads to more reliable decision-making \cite{yang20243wc}.

\subsection{Twin Support Vector Regression}
The twin support vector regression (TSVR) model formulates the regression task by solving two smaller and independent quadratic programming problems. The first optimization problem learns the parameters of the lower bounding regressor and is given by
\begin{align}
   \underset{w_1, b_1}{\min} & ~ \frac{1}{2} \| Y - e\epsilon_1 - (Aw_1 + eb_1)\|^2 + \mathcal{D}_1 e^T\eta_1 \nonumber \\
  \text{s.t.} &~~ Y - e\epsilon_1 - (Aw_1 + eb_1) \geq -\eta_1, \nonumber \\
 & ~~\eta_1 \geq 0,
\end{align}
whereas the second optimization problem determines the parameters of the upper bounding regressor and is expressed as
\begin{align}
   \underset{w_2, b_2}{\min} & ~ \frac{1}{2} \| Y + e\epsilon_2 - (Aw_2 + eb_2)\|^2 + \mathcal{D}_2 e^T\eta_2 \nonumber \\
  \text{s.t.} &~~  (Aw_1 + eb_1) - Y - e\epsilon_2 \geq -\eta_2, \nonumber \\
 & ~~\eta_2 \geq 0,
\end{align}
where $\mathcal{D}_1,~ \mathcal{D}_2$ are regularization parameters, $\epsilon_1,~\epsilon_2 > 0$ define the insensitive zones, $\eta_1$ and $\eta_2$ denote the slack variable vectors.

\renewcommand{\thesection}{S.2}
\section{Experimental setup}
The experiments were conducted on a workstation powered by an Intel(R) Xeon(R) Gold 6226R processor operating at 2.90 GHz, paired with 256 GB of RAM. The system runs Windows 11, and all implementations were carried out in Python 3.11. Each dataset was randomly split into training and testing sets using a 70/30 proportion. To determine the optimal hyperparameters, we employed five-fold cross-validation together with grid search across the range $\mathcal{D}_1 = \mathcal{D}_2 = \{10^{-5}, 10^{-4}, \ldots, 10^5\}$. The parameters $\epsilon_1$ and $\epsilon_2$ are tuned by selecting values from the set \\
$\{0.001, 0.005, 0.01, 0.05, 0.1, 0.15, 0.2, 0.25\}$. To assess the performance of the proposed GBTSVR and GB-TSVQR models, we employ four metrics, including mean absolute error (MAE), root mean squared error (RMSE), negative error (Neg Error), and positive error (Pos Error). The following are the specific definitions of these metrics:
\begin{align}
   & \text{RMSE} = \sqrt{\frac{1}{n} \sum_{i=1}^n (f(x_i) - y_i)^2}, \\
    & \text{MAE} = \frac{1}{n} \sum_{i=1}^n \lvert (f(x_i) - y_i)\rvert, 
\end{align}
\begin{align}
    & \text{Pos Error} = \frac{1}{n}\sum_{i=1, f(x_i)\leq y_i }^n  \lvert (f(x_i) - y_i)\rvert, \\
    & \text{Neg Error} = \frac{1}{n}\sum_{i=1, f(x_i) > y_i }^n  \lvert (f(x_i) - y_i)\rvert.
\end{align}

\renewcommand{\thesection}{S.3}
\section{Experiments on Time-series Datasets}

\renewcommand{\thetable}{S.1}
\begin{table*}[ht!]
\centering
    \caption{Performance comparison of the proposed GBTSVR and GB-TSVQR models and baseline models on time-series datasets in terms of RMSE, MAE, Poss Error, and Neg Error.}
    \label{performance time-series datasets}
    \resizebox{1\linewidth}{!}{
\begin{tabular}{llccccccc}
\hline
Dataset & Metric & SVR \cite{basak2007support} & TSVR \cite{peng2010tsvr} & TSVQR \cite{ye2024twin} & $L_1$-TSVQR \cite{ye2025l1} & GL-TPSVR \cite{liu2026twin} & GBTSVR$^{\dagger}$ & GB-TSVQR$^{\dagger}$ \\  \hline
NNGC1\_dataset\_D1\_V1\_003 & RMSE & 0.9938 & 0.01041 & 0.81541 & 0.00875 & 0.98186 & 0.00064 & 0.00073 \\
$(430, 5)$ & MAE & 0.84253 & 0.00826 & 0.68345 & 0.00786 & 0.83049 & 0.00053 & 0.00072 \\
 & Pos Error & 0.76537 & 0.00662 & 0.65806 & 0.00736 & 0.86224 & 0.00059 & - \\
 & Neg Error & 0.90556 & 0.01014 & 0.70846 & 0.00742 & 0.802 & 0.00048 & 0.00072 \\ \hline
NNGC1\_dataset\_D1\_V1\_004 & RMSE & 0.65551 & 0.00277 & 0.46681 & 0.0084 & 0.65176 & 0.00033 & 0.00085 \\
$(545, 5)$ & MAE & 0.54944 & 0.00241 & 0.35603 & 0.0075 & 0.5423 & 0.00026 & 0.00077 \\
 & Pos Error & 0.38007 & 0.00191 & 0.38925 & 0.0073 & 0.37873 & 0.00031 & 0.00022 \\
 & Neg Error & 0.60589 & 0.00264 & 0.29995 & 0.0065 & 0.60606 & 0.0002 & 0.00008 \\ \hline
NNGC1\_dataset\_D1\_V1\_005 & RMSE & 0.04651 & 0.00061 & 0.2303 & 0.00354 & 0.41303 & 0.00892 & 0.00011 \\
$(430, 5)$ & MAE & 0.03901 & 0.00051 & 0.2275 & 0.00141 & 0.41152 & 0.00726 & 0.00009 \\
 & Pos Error & 0.04316 & 0.00053 & - & 0.00041 & 0.41152 & 0.00805 & - \\
 & Neg Error & 0.03287 & 0.0005 & 0.2275 & 0.00082 & - & 0.00466 & 0.00001 \\ \hline
NNGC1\_dataset\_D1\_V1\_006 & RMSE & 0.25026 & 0.00862 & 0.33471 & 0.00203 & 0.47458 & 0.00194 & 0.00047 \\
$(610, 5)$ & MAE & 0.20081 & 0.00515 & 0.28032 & 0.00147 & 0.41319 & 0.00157 & 0.00045 \\
 & Pos Error & 0.19707 & 0.0035 & 0.15724 & 0.00203 & 0.42924 & 0.00149 & 0.00045 \\
 & Neg Error & 0.20309 & 0.00783 & 0.3064 & 0.00776 & 0.11591 & 0.00171 & 0.0001 \\ \hline
NNGC1\_dataset\_D1\_V1\_007 & RMSE & 0.10424 & 0.00311 & 0.19239 & 0.00885 & 0.4195 & 0.00405 & 0.00054 \\
$(610, 5)$ & MAE & 0.08292 & 0.00196 & 0.16827 & 0.00418 & 0.40649 & 0.00325 & 0.00052 \\
 & Pos Error & 0.09092 & 0.00122 & 0.07523 & 0.00102 & 0.40649 & 0.00348 & 0.00052 \\
 & Neg Error & 0.07657 & 0.00316 & 0.17842 & 0.00238 & - & 0.00287 & - \\ \hline
NNGC1\_dataset\_D1\_V1\_008 & RMSE & 0.10357 & 0.00411 & 0.2791 & 0.00106 & 0.42315 & 0.00368 & 0.00011 \\
$(540, 5)$ & MAE & 0.07853 & 0.00305 & 0.25947 & 0.00087 & 0.41017 & 0.00293 & 0.00008 \\
 & Pos Error & 0.07673 & 0.00288 & 0.03178 & 0.00109 & 0.41272 & 0.00303 & 0.00007 \\
 & Neg Error & 0.08046 & 0.00361 & 0.26088 & 0.00604 & 0.00067 & 0.0027 & 0.00095 \\ \hline
NNGC1\_dataset\_D1\_V1\_009 & RMSE & 0.55713 & 0.03569 & 0.56568 & 0.00306 & 0.6612 & 0.00253 & 0.00038 \\
$(540, 5)$ & MAE & 0.43811 & 0.01896 & 0.4437 & 0.00101 & 0.53367 & 0.0019 & 0.00031 \\
 & Pos Error & 0.44102 & 0.01095 & 0.43426 & 0.00306 & 0.58708 & 0.00143 & 0.0001 \\
 & Neg Error & 0.43566 & 0.03257 & 0.45002 & 0.00829 & 0.36526 & 0.00228 & 0.00034 \\ \hline
NNGC1\_dataset\_D1\_V1\_010 & RMSE & 0.12526 & 0.0058 & 0.36128 & 0.00919 & 0.4219 & 0.00441 & 0.00045 \\
$(585, 5)$ & MAE & 0.10006 & 0.00319 & 0.33974 & 0.00418 & 0.40322 & 0.00345 & 0.00042 \\
 & Pos Error & 0.09756 & 0.00173 & 0.01862 & 0.0012 & 0.40322 & 0.0033 & 0.00004 \\
 & Neg Error & 0.10261 & 0.0052 & 0.34158 & 0.00274 & - & 0.0037 & 0.0003 \\ \hline
NNGC1\_dataset\_E1\_V1\_008 & RMSE & 0.24381 & 0.00426 & 0.39578 & 0.00239 & 0.46686 & 0.00133 & 0.0006 \\
$(740, 5)$ & MAE & 0.20354 & 0.00304 & 0.31672 & 0.00168 & 0.41277 & 0.00115 & 0.00058 \\
 & Pos Error & 0.16215 & 0.0031 & 0.1089 & 0.00247 & 0.4531 & 0.00126 & 0.00006 \\
 & Neg Error & 0.27285 & 0.00293 & 0.32652 & 0.00619 & 0.0638 & 0.00093 & - \\ \hline
NNGC1\_dataset\_E1\_V1\_009 & RMSE & 0.23091 & 0.00517 & 0.40271 & 0.00227 & 0.46194 & 0.00147 & 0.00046 \\
$(740, 5)$ & MAE & 0.19294 & 0.00391 & 0.33089 & 0.00163 & 0.41035 & 0.00126 & 0.00044 \\
 & Pos Error & 0.151 & 0.00381 & 0.10054 & 0.00237 & 0.44534 & 0.00141 & 0.00001 \\
 & Neg Error & 0.26455 & 0.00421 & 0.3395 & 0.007 & 0.05693 & 0.00097 & 0.00045 \\ \hline
NNGC1\_dataset\_E1\_V1\_010 & RMSE & 0.44161 & 0.00815 & 0.48839 & 0.0033 & 0.60468 & 0.0012 & 0.00011 \\
$(650, 5)$ & MAE & 0.35476 & 0.00544 & 0.38271 & 0.00131 & 0.49549 & 0.00093 & 0.00009 \\
 & Pos Error & 0.35456 & 0.0055 & 0.39629 & 0.00031 & 0.5912 & 0.00093 & 0.00023 \\
 & Neg Error & 0.35498 & 0.00537 & 0.37188 & 0.00877 & 0.20837 & 0.00092 & 0.00092 \\ \hline
Average RMSE &  & 0.34115 & 0.00801 & 0.41205 & 0.0048 & 0.54368 & 0.00277 & 0.00044 \\ \hline
Average Rank &  & 5.36 & 3.45 & 5.82 & 3.18 & 6.82 & 2.18 & 1.18 \\ \hline
\multicolumn{9}{l}{The proposed models is denoted by $^{\dagger}$.}
\end{tabular}}
\end{table*}

To demonstrate the effectiveness of the proposed GBTSVR and GB-TSVQR models in real-world regression scenarios, we conduct extensive experiments on benchmark time-series datasets. The comparative results of the proposed models against several state-of-the-art baseline methods are reported in Table \ref{performance time-series datasets}, using RMSE, MAE, Pos Error, and Neg Error as evaluation metrics. From the results, it is evident that the proposed granular-ball-based models consistently deliver superior predictive performance across the considered time-series datasets. In particular, GB-TSVQR achieves the best RMSE performance on the majority of datasets, closely followed by GBTSVR, significantly outperforming classical SVR variants and other twin-based regression models. The average RMSE values further validate this observation. GB-TSVQR attains the lowest average RMSE among all competing models, while GBTSVR also achieves a notably lower average RMSE compared to the baseline models. This indicates that incorporating granular-ball computing effectively enhances the robustness and generalization capability of TSVR frameworks, particularly for complex and noisy time-series data. The consistent dominance of GBTSVR and GB-TSVQR across multiple metrics highlights their strong modeling capacity and robustness. These results confirm that the proposed GBTSVR and GB-TSVQR models are well suited for time-series regression tasks and provide a powerful alternative to existing SVR-based and twin regression models.

\end{document}